\pdfoutput=1

\documentclass[11pt]{article}

\usepackage{acl}

\usepackage{times}
\usepackage{latexsym}

\usepackage[T1]{fontenc}

\usepackage[utf8]{inputenc}

\usepackage{microtype}

\usepackage[pdftex]{graphicx}
\usepackage{amsfonts,amsmath}
\usepackage{booktabs}
\usepackage{float}
\usepackage[ruled,vlined,noend]{algorithm2e}
\SetKwComment{Comment}{$\triangleright$\ }{}

\title{Towards a Progression-Aware Autonomous Dialogue Agent}

\author{Abraham Sanders$^1$, Tomek Strzalkowski$^1$, Mei Si$^1$, Albert Chang$^1$, \\
        \bf{Deepanshu Dey}$^1$, \bf{Jonas Braasch}$^1$, \bf{Dakuo Wang}$^2$\\
        $^1$Rensselaer Polytechnic Institute, Troy, NY, USA, $^2$ IBM Research, USA \\
        \texttt{\{sandea5,tomek,sim,changa4,deyd,braasj\}@rpi.edu}\\
        \texttt{dakuo.wang@ibm.com}
        }

\begin{document}
\maketitle
\begin{abstract}
Recent advances in large-scale language modeling and generation have enabled
the creation of dialogue agents that exhibit human-like responses in a wide
range of conversational scenarios spanning a diverse set of tasks, from 
general chit-chat to focused goal-oriented discourse. While these agents
excel at generating high-quality responses that are relevant to prior context, they suffer from a lack of awareness of the overall direction in which the conversation is headed, and the likelihood of task success inherent therein. Thus, we propose a framework in which dialogue agents can evaluate the progression of a conversation toward or away from desired outcomes, and use this signal to inform planning for subsequent responses. Our framework is composed of three key elements: (1) the notion of a "global" dialogue state (GDS) space, (2) a task-specific progression function (PF) computed in terms of a conversation's trajectory through this space, and (3) a planning mechanism based on dialogue rollouts by which an agent may use progression signals to select its next response.
\end{abstract}

\section{Introduction}
All human conversation serves some purpose. These may range from negotiating an agreement to explaining a topic to maintaining a social relationship. People are generally capable of forming an assessment, sometimes subconsciously, whether a conversation is going well or not and adjusting their behavior accordingly. Such assessment, which underlies most human conversation, is essential in continuous awareness of the direction where the interaction is heading and whether the parties are in sync or not, e.g., \citet{Ber:91}. In a task-oriented interaction, the participants assess if progress towards a successful outcome is being made. In a negotiation, parties assess if an agreement is likely. Even in a casual conversation,
\begin{figure}[H]
\vskip 0.05in
\begin{center}
\hspace*{-0.5cm} 
\includegraphics[width=1.05\columnwidth]{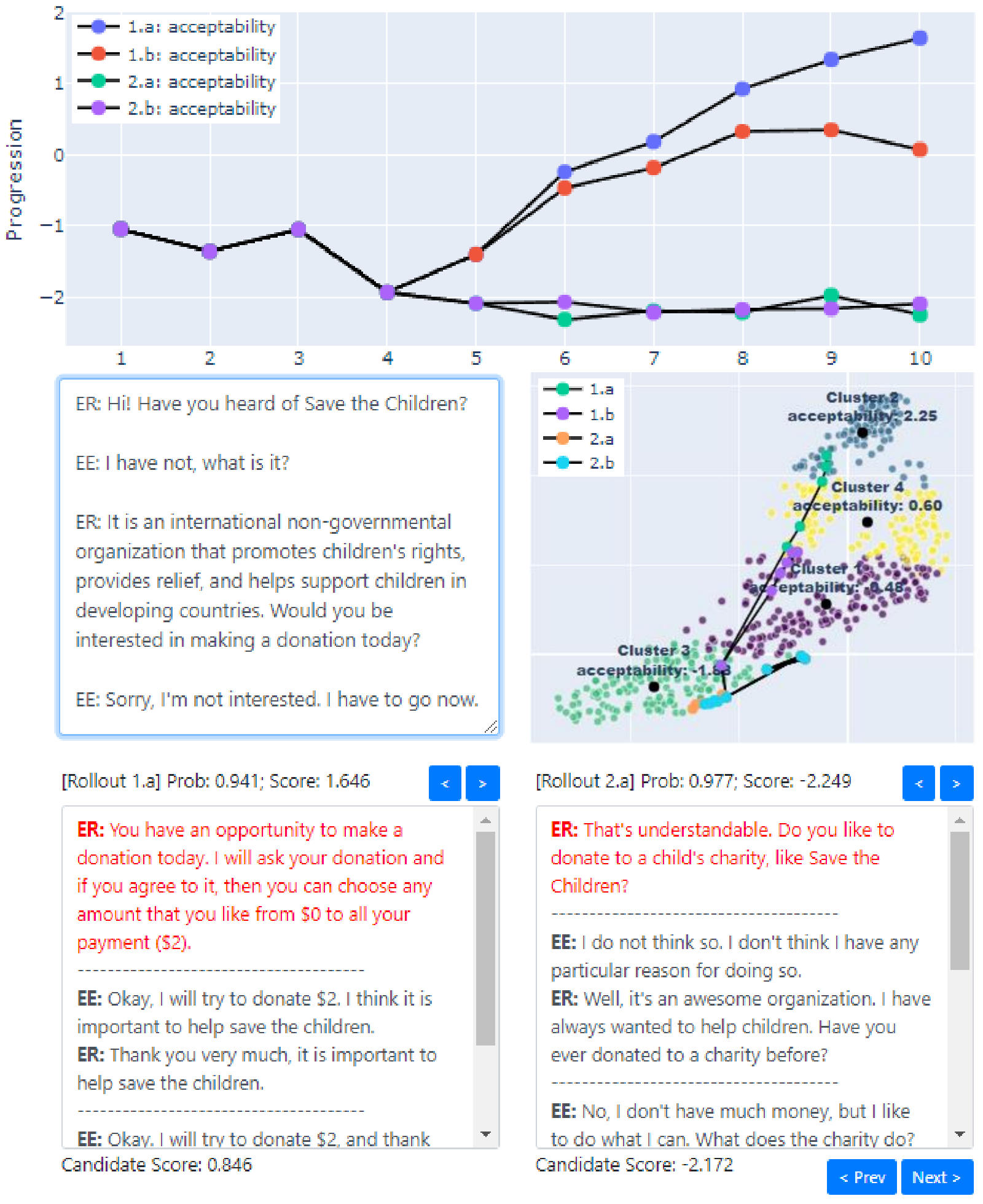}
\caption{Our framework applied to the charity solicitation task in Persuasion For Good \citep{Wan:19}. Given the dialogue history (center left), the system uses rollouts \citep{lewis-etal-2017-deal} to simulate the outcome of two response candidates (bottom, in red). Each rollout is mapped as a path through the Global Dialogue State space (center right) where it can be compared with similar outcomes. The candidates are finally ranked using the Progression Function (top), and the best is selected.}
\label{demo}
\end{center}
\vskip -0.2in
\end{figure}
\noindent people intuitively sense when to continue, when to change the subject, or when to stop. Based on such (subjective) assessment, participants adjust what to say next: whether to push forward, make a concession, soften the tone, digress, or say goodbye. A wide range of research in conversation and discourse analysis is devoted to these and related issues including \citep{Bee:20, Cas:07, Fri:04, Gre:08, Lan:03}; however, recent efforts in Dialogue State Tracking (DST) have been primarily focused on collecting fine-grained details (e.g., slot-value pairs for travel booking or restaurant reservation) \citep{balaraman2021recent} without concern for the overall direction and quality of the conversation, even though the latter is critical for achieving human-level dialogue interaction.

As such, we approach dialogue state tracking at a higher level, focusing instead on what we call the \textbf{Global Dialogue State (GDS)}. Given a conversational task (e.g., negotiation), the global state of a dialogue reflects the most likely outcome (e.g., a strong agreement or a stalemate) given the history of the dialogue up to the current turn. In contrast to traditional DST, the global state remains invariant to the specific details discussed at each turn (e.g., names, dates, quantities) that are typically the concern of slot-filling models. Rather, global dialogue states are influenced by the contexts in which these details occur (e.g., ``I \textit{would love to} donate \$5 to this charity!'' vs. ``I \textit{would never} donate \$5 to this charity''). Thus, the global state of a dialogue can be measured in terms of its \textit{semantic similarity} to other groups of dialogues for the same task, which can be naturally formulated as a cluster-assignment problem in the dialogue embedding space. That is, a dialogue which is assigned at the current turn to a cluster of highly successful outcomes may assume a high likelihood of success, and likewise a dialogue assigned to a cluster of unsuccessful outcomes may assume a low likelihood of success. It follows from this that the path of a dialogue through global state space can be used to derive a \textbf{Progression Function (PF)} to give turn-level estimates of task success, which can in turn be used by a dialogue agent to inform its next response.

The remainder of this paper is organized as follows: In Section \ref{section_related_work} we review relevant literature pertaining to dialogue state tracking and response planning; in Section \ref{section_methods} we formally define the global dialogue state and progression function, propose \textbf{supervised} and \textbf{unsupervised} approaches for modeling them, and describe how they can be used to assess and select dialogue response candidates; in Section \ref{section_experiments} we experimentally apply our framework to the charity solicitation task in the Persuasion For Good dataset \citep{Wan:19}, reporting results from automatic and manual evaluations; and in Sections \ref{section_limitations_future_directions} and \ref{section_conclusion} we conclude with a discussion of limitations and future directions. Code for our methods and experiments has been released, \footnote{Available at \url{https://github.rpi.edu/LACAI/dialogue-progression}} 
and a listing of software packages we use can be found in Appendix \ref{section_software_packages_used}.

\section{Related Work} \label{section_related_work}
Our work lies at the intersection of dialogue state tracking and response planning. As previously noted, we approach dialogue state at a much higher level than is typically seen in the DST literature. Our concept of global dialogue state is not mutually exclusive with traditional DST approaches, which we refer to from here on as \textbf{local} DST. Rather, an effective dialogue system might integrate local and global DST approaches to enable simultaneous tracking of user intents and slot-value pairs (needed for interfacing with external resources) and the overall likelihood of conversational success.

\subsection{Dialogue State Tracking} \label{subsection_dialogue_state_tracking}
Local DST approaches are used in task-oriented (also called goal-oriented) dialogue systems. Local DST is responsible for identifying user intent (e.g., search for restaurants) and extracting slot-value pairs (e.g., location, price range). Recent DST systems perform state tracking in a diverse set of domains, including food ordering \citep{Ler:21}, travel reservations \citep{Gin:21}, negotiations \citep{he-etal-2018-decoupling}, and many others. Datasets such as MultiWOZ \citep{budzianowski-etal-2018-multiwoz, eric-EtAl:2020:LREC, zang-etal-2020-multiwoz} and SGD \citep{Rastogi_Zang_Sunkara_Gupta_Khaitan_2020} provide large-scale testbeds for training single DST systems that generalize across many task domains. However, local DST is generally not deployed in open-domain end-to-end dialogue systems that focus on social interaction and user engagement, recent examples including DialoGPT \citep{zhang-etal-2020-dialogpt}, Meena \citep{adiwardana2020towards}, and BlenderBot \citep{roller-etal-2021-recipes, xu2021beyond}. In open-domain models, the task is unconstrained and thus it makes little sense to employ traditional slot-based dialogue state trackers. Instead, these models track state implicitly in their latent representations of dialogue history. Unlike local DST, global state tracking is applicable in both the task-oriented and open-domain settings.

\subsection{Dialogue Response Planning}
Many approaches exist for planning in dialogue response generation. Planning helps a dialogue agent maintain coherence over multiple turns and stay on track to complete its goal. \citet{lewis-etal-2017-deal} introduce Dialogue Rollouts, allowing a negotiation agent to simulate the remainder of a conversation based on each of multiple candidate responses and select the one which yields the best outcome. \citet{pmlr-v80-yarats18a} follow up by separating semantic planning and surface realization for response generation by first producing a latent semantic representation of the dialogue plan and then conditioning on it during generation with rollouts. Similarly, \citet{jiang-etal-2019-towards} implement a look-ahead module to implicitly predict multiple future turns in an end-to-end encoder-decoder architecture, experimenting with negotiation and restaurant reservation settings. These works all experiment in task domains where goal achievement is explicitly measurable, which is not true in the general case. Thus we propose to combine such methods with our progression function which provides estimates of goal completion likelihood. Particularly, in this paper we demonstrate the use of rollouts with the PF as a reward signal.

\section{Methods} \label{section_methods}
\begin{figure*}[t]
\vskip -0.20in
\begin{center}
\includegraphics[width=\textwidth]{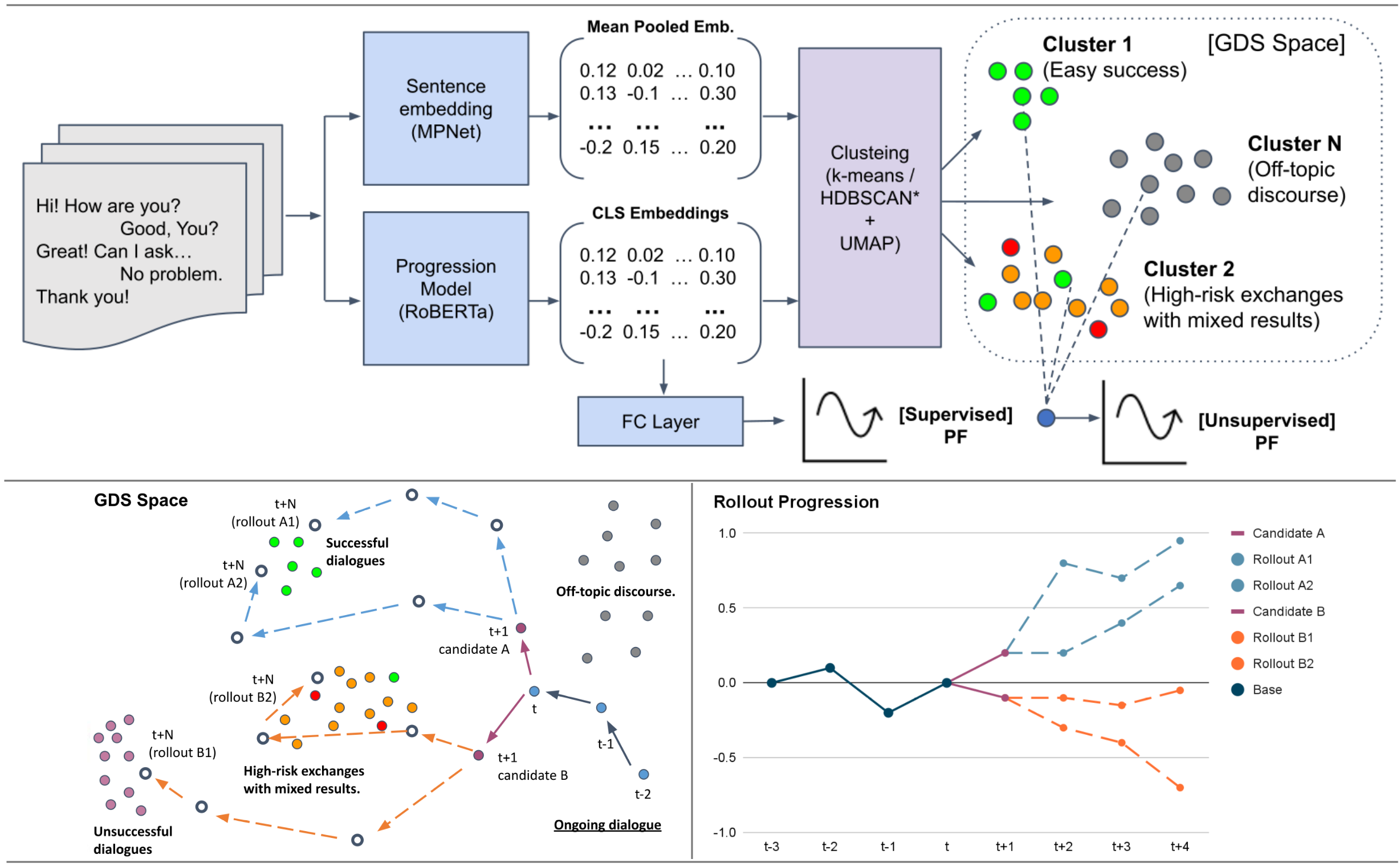}
\caption{Architecture of the supervised and unsupervised GDS and PF models (top). In GDS space (top right), each cluster is characterized by similar dialogue semantics, and is thus interpreted as the class of typical outcomes for dialogues within. GDS and PF can be used with rollouts (bottom) to allow a dialogue agent to plan ahead. }
\label{architecture}
\end{center}
\vskip -0.2in
\end{figure*} 

The goal of our system is to construct a global dialogue state space for a task-specific dataset and learn a progression function to estimate how well an ongoing dialogue is progressing toward the desired outcome of the task. The quantity output by the progression function is an estimate of a dialogue-level attribute which indicates task success  (e.g. satisfaction in a customer service task). In many task domains, the success of a conversation cannot be completely measured by a single attribute. For example, in the charity solicitation task we use in our experiments, donation amount is the primary success attribute. Here, there are cases where the conversation appears to go very well, but ultimately no donation is made for unexpected reasons such as the solicitee not being able to afford to donate. One could reasonably expect such an outcome to be ``acceptable'' in the context of a solicitation task since the solicitee has engaged with the solicitor and displayed interest, and we cannot reasonably expect the solicitor to force a donation out of someone who cannot afford it. Thus we introduce the ``\textbf{acceptability score}'', a synthetic attribute that measures success by considering multiple factors (e.g., donation amount and sentiment). For any dialogue dataset, the acceptability score combines multiple dialogue-level attributes in a way sensitive to their covariance with the primary success attribute:
\begin{align}
    \text{ACC}_D = \text{prim}_D + \sum_{i=1}^{|\mathbf{v}_D|}\text{Cov}(\text{prim},  \text{attr}_i)\cdot {\mathbf{v}_D}_i
\end{align}
where $\text{prim}_D$ is the primary success attribute (e.g. donation amount) value for dialogue $D$, $\mathbf{v}_D$ is the vector of all other attribute values (e.g., sentiment) for dialogue $D$, and $\text{Cov}(\text{prim}, \text{attr}_i)$ is the training set covariance between the primary success indicator and the $i$-th other attribute. We define the output of the progression function to be an estimate of the acceptability score.

To learn the progression function, dialogue-level attribute annotations must exist for use in this purpose. However, in many settings such annotations are not available in sufficient quantity to directly learn a progression model with sufficient generalization. Consequently, we propose \textbf{supervised} and \textbf{unsupervised} approaches for learning the global state and progression models.

\subsection{Unsupervised Approach} \label{subsection_unsupervised_approach}
\subsubsection{Global Dialogue State}\label{subsubsection_global_dialogue_state}
In the unsupervised approach, the GDS space is a dialogue embedding space where clusters of embeddings represent groups of dialogues with similar semantic content. For each complete dialogue $D$ in the training set, all utterances are independently embedded and then pooled to create a dialogue-level embedding $\mathbf{u}_D \in \mathbb{R}^d$ where $d$ is the embedding size. The GDS space is thus given as a matrix in $\mathbb{R}^{N\times d}$ where $N$ is the number of complete dialogues. To embed utterances we take advantage of pre-trained sentence encoders exposed to large-scale corpora. Specifically, we use a publicly available MPNet \citep{NEURIPS2020_c3a690be} model fine-tuned for semantic textual similarity using a contrastive objective on over 1B training pairs from 32 distinct datasets. \footnote{Obtained from \url{https://huggingface.co/sentence-transformers/all-mpnet-base-v2}}
To combine utterance embeddings into a dialogue-level embedding we use recency-weighted mean pooling. The recency weight $\beta$ determines how much emphasis is placed on more recent utterances, where $\beta=0$ means all utterances are weighted evenly and $\beta > 0$ means that more emphasis is placed on more recent utterances. The motivation for recency weighting is to test the hypothesis that more recent developments in a conversation are more relevant for predicting current progression toward a goal. For example, a conversation may start out off-task with participants engaging in small talk, and then later re-focus.

\begin{figure}[h]
\vskip 0.05in
\begin{center}
\includegraphics[width=\columnwidth]{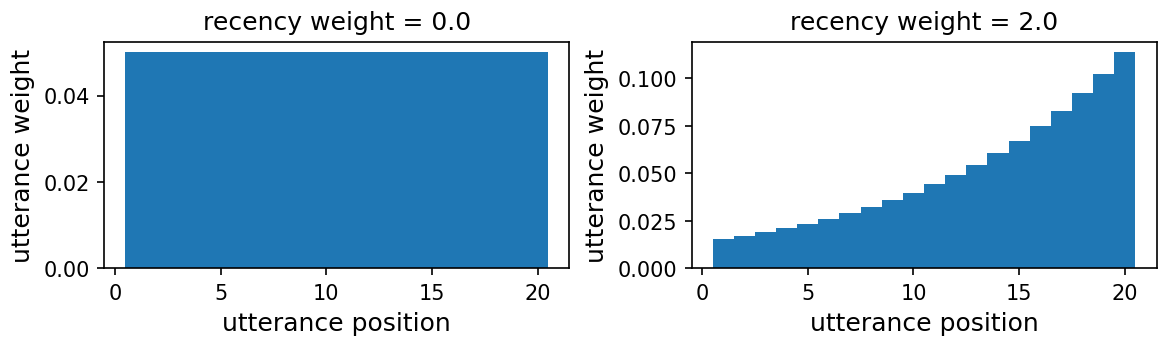}
\caption{Recency weight $\beta$ controls how much emphasis is placed on recent utterances when computing $\mathbf{u}_D$. }
\label{recency-weighting}
\end{center}
\vskip -0.2in
\end{figure}

The embedding for dialogue $D$ with $|D|$ utterances is thus formulated as $\mathbf{u}_D = U^T\text{softmax}(\mathbf{r})$
where $U$ is the matrix of utterance vectors in $\mathbb{R}^{|D|\times d}$ and $\mathbf{r} \in \mathbb{R}^{|D|}$ is a vector of evenly spaced real numbers over the interval $[0, \beta]$. The softmax ensures all recency weights sum to 1 and can be interpreted as probabilities as done with attention scores in \citep{bahdanau2014neural, vaswani2017attention}. As shown in Figure \ref{recency-weighting}, each utterance is thus weighted by a monotonically increasing probability mass where higher values of $\beta$ cause more mass to be concentrated at the end of the dialogue.

The unsupervised GDS model is a clustering of the dialogues in their embedding space. The dialogue embeddings are either clustered directly or after projection to a lower-dimensional space using Parametric UMAP \citep{sainburg2020parametric, mcinnes2018umap}. We experiment with k-means and HDBSCAN \citep{mcinnes2017accelerated, campello2013density} to cluster the embeddings. For k-means, we choose the number of clusters $k$ and train with 10 random initializations. For HDBSCAN, we choose the minimum cluster size and minimum samples hyperparameters, and the optimal number of clusters are selected automatically. Unlike k-means which simply partitions the embedding space, HDBSCAN classifies some embeddings as noise points. Clustering hyperparameters are selected by cross-validation on several metrics as described later in Section \ref{section_experiments}. The process of constructing the GDS model is illustrated in Figure \ref{architecture}.

The clusters output by this process can be interpreted as the equivalence classes of final global states possible for the task represented in the dialogue dataset. To estimate the global state of an ongoing dialogue $D^\prime$, it is embedded as $\mathbf{u}_{D^\prime} \in \mathbb{R}^d$ in the same manner as the complete training dialogues, followed by optional dimensionality reduction. The trained k-means or HDBSCAN model is then used to assign $D^\prime$ to one of the existing clusters, or possibly as a noise point in the case of HDBSCAN.

Each cluster is assigned an aggregate acceptability score by taking an average of acceptability for each dialogue in the cluster. If k-means is used, we aggregate using a 10\% trimmed mean across all dialogues in the cluster. If HDBSCAN is used, a probability is returned for each dialogue representing the likelihood that it is a member of its assigned cluster, so we compute the probability-weighted average across all dialogues in the cluster. Dialogues classified as noise points are ignored.

To visualize the GDS model, Parametric UMAP is used again to project the clustered dialogue embeddings into $\mathbb{R}^2$ or $\mathbb{R}^3$. As shown in Figure \ref{demo}, the GDS model can be mapped as a scatter plot with each cluster labeled by its aggregate values. If k-means is used, the cluster centroids can be displayed as a bold point within each cluster. HDBSCAN clusters do not have centroids, but they do have a number of representative points that are close to the cluster core. We average these points to simulate a centroid for display purposes, and likewise show it as a bold point within each cluster. To show how an ongoing dialogue $D^\prime$ traverses the GDS space over time, its embeddings at each turn $t$ are projected onto the map and connected with line segments to form a path.

\subsubsection{Computing Progression}\label{subsubsection_computing_progression}
Since each cluster in the GDS space is intended to represent a class of end-task global states, we compute the progression of an ongoing dialogue $D^\prime$ with respect to the likelihood that its final global state will rest in each individual cluster. Supposing there are $k$ final clusters after running k-means or HDBSCAN, we compute a probability vector $\mathbf{p}_{D^\prime} \in \mathbb{R}^k$ such that 
${\mathbf{p}_{D^\prime}}_i = P(\mathbf{u}_{D^\prime} \in C_i)\ \text{for}\ i \in \{1, \dots k\}$
where $C_i$ is cluster $i$. $\mathbf{p}_{D^\prime}$ is computed differently for k-means and HDBSCAN. K-means does not produce a probabilistic soft clustering, so we define $\mathbf{p}_{D^\prime}$ with respect to the proximity of $\mathbf{u}_{D^\prime}$ to the centroids of each cluster:
\begin{align}
    \mathbf{p}_{D^\prime} = \text{softmax}\left(\frac{1}{||\mathbf{u}_{D^\prime}-\mathbf{c}_i||_2}\ :\ i \in \{1, \dots k\} \right)
\end{align}
where $\textbf{c}_i \in \mathbb{R}^d$ is the centroid of cluster $i$.
HDBSCAN does produce a probabilistic soft clustering, so in that case $\mathbf{p}_{D^\prime}$ is already computed.

We ultimately want the closest (or most probable) clusters for ongoing dialogue $D^\prime$ to have the most sway in estimating its progression at the current point in time. That is, if $D^\prime$ has moved into a cluster of high-success outcomes, its progression should increase. Likewise if $D^\prime$ has moved away from such a high-success cluster, either into a lower-success cluster or off-task into a noisy or unknown region of the GDS space, its progression should decrease. Thus, once $\mathbf{u}_{D^\prime}$ is computed, we estimate its progression as the probability-weighted average of the aggregate acceptability scores assigned to each cluster. This is formulated as
\begin{align}
    \text{PROG}(\mathbf{u}_{D^\prime}) = \frac{\mathbf{v}^T\mathbf{p}_{D^\prime}}{\sum_{i=1}^k{\mathbf{p}_{D^\prime}}_i}
\end{align}
where $\mathbf{v} \in \mathbb{R}^k$ is a vector of the aggregate acceptability scores assigned to each cluster. The scaling factor in the denominator ensures that ongoing dialogue embeddings classified as noise points by HDBSCAN will not be assigned progression values close to zero as a consequence of not belonging to any cluster, which can cause significant fluctuation in the progression function as the dialogue traverses noisy regions of the GDS space. \footnote{For HDBSCAN we also experiment with softmax for re-scaling $\mathbf{v}^T\mathbf{p}_{D^\prime}$, giving $\text{PROG}(\mathbf{u}_{D^\prime}) = \text{softmax}(\mathbf{v}^T\mathbf{p}_{D^\prime})$.} Figure \ref{architecture} illustrates how progression of an ongoing dialogue depends on its position in GDS space.

\subsection{Supervised Approach} \label{subsection_supervised_approach}
For the supervised approach, we simply fine-tune RoBERTa \citep{liu2019roberta} to directly predict acceptability given the dialogue history text, where all utterances are concatenated into a single sequence. To construct the GDS space we obtain the dialogue level embedding $\mathbf{u}_{D}$ directly from the CLS (<s>) token for each complete dialogue in the training set, and cluster them as in Section \ref{subsubsection_global_dialogue_state}. Unlike the unsupervised approach where recency weighting is used to ``attend'' to more recent parts of the dialogue, the supervised fine-tuning process causes the CLS embedding to aggregate the parts of the dialogue most relevant to the task objective, which is more optimal than the recency heuristic. Also, unlike the unsupervised approach where progression for an ongoing dialogue is computed with respect to its embedding, here progression is directly predicted by RoBERTa. In our experiments we compare RoBERTa-base, RoBERTa-large, and RoBERTa-large-adapted, the latter receiving additional domain adaptation training for dialogue. Domain adaptation is done via Masked Language Modeling (MLM) on a self-generated version of the Gutenberg Dialogue Dataset \citep{csaky-recski-2021-gutenberg}. Hyperparameters and model weights from domain adaptation training are provided with our code release.

\subsection{Response Planning} \label{subsection_response_planning}
To allow a dialogue agent to use the progression function as feedback for response planning, we adopt \textbf{Dialogue Rollouts} \citep{lewis-etal-2017-deal} to simulate the outcomes of a set of response candidates. A rollout for a response candidate simulates the next $N$ turns of the conversation (for both participants) given that candidate is used. At each turn of a negotiation task, \citet{lewis-etal-2017-deal} sample a set of $c$ response candidates and $s$ rollouts per candidate. They score each rollout by a deterministic reward (the value of the items ``won'' by the agent during negotiation), and rank each candidate by the average of its rollout scores. The highest ranking candidate is then selected by the agent. As shown in Figure \ref{architecture}, we generalize this process to any task for which a progression function can be learned, replacing the deterministic reward with the progression function value. To demonstrate this, we fine-tune the 762M parameter DialoGPT \citep{zhang-etal-2020-dialogpt} \footnote{Obtained from \url{https://huggingface.co/microsoft/DialoGPT-large}} as a dialogue response generator and use beam sampling to generate response candidates and rollouts. We select DialoGPT for this task as it is pre-trained on a large Reddit dialogue corpus.

\section{Experiments} \label{section_experiments}
\subsection{Dataset} \label{subsection_dataset}
We apply our framework to the Persuasion For Good dataset \citep{Wan:19}, which is a crowd-sourced dialogue dataset where the task for an individual playing the role of persuader is to convince another individual playing the role of persuadee to make a donation to a well-known children's charity. We selected this dataset since it has a clear task objective (to solicit donations), but a complex relationship between dialogue content and success. The dataset authors identify 10 distinct persuasion strategies used to solicit donations, where different strategies correlate with donation amount at different strengths. Additionally, participants in Persuasion For Good dialogues complete a pre-task psychological survey, yielding 23 attributes based on the Big-Five personality traits \citep{goldberg1992development}, the Moral Foundations endorsement \citep{graham2011mapping}, the Schwartz Portrait Value \citep{cieciuch2012comparison}, and the Decision-Making style \citep{hamilton2016development} questionnaires for each individual. The dataset authors demonstrated varying degrees of correlation between these psychological attributes and the end-task donation amount. The complexity in measuring progression in this context, coupled with it being a relatively small dataset, makes Persuasion For Good an interesting and challenging testbed for our framework. Persuasion For Good contains 1017 dialogues, each with approximately 10 turns per speaker (20 utterances). 

\subsection{Progression Function Experiments} 
\label{subsection_progression_function_experiments}
As the objective of the task is to solicit donations, we consider the end-dialogue persuadee donation amount to be the primary dialogue success indicator. We also augment the dataset by computing average dialogue sentiment. To compute sentiment we use a RoBERTa model\footnote{Obtained from \url{https://huggingface.co/cardiffnlp/twitter-roberta-base-sentiment}} fine-tuned on the sentiment classification task of the TweetEval benchmark \citep{barbieri-etal-2020-tweeteval}, which was publicly released by the benchmark authors. We score sentiment at the utterance level in the range $[-1, 1]$ by multiplying the sentiment class probabilities predicted by RoBERTa for negative, neutral and positive by $\{-1, 0, 1\}$ respectively and summing the result. We then average the utterance-level sentiment score for each dialogue.

We filter the dataset to eliminate dialogues with end-task donation amounts outside the allowed task parameters (between \$0 and \$2 USD), and use a regular expression to filter out dialogues where the persuadee fails to make a donation after promising a non-zero dollar amount in the conversation. After filtration we are left with 751 dialogues for our study. We split the dialogues into a training and test set, leaving 577 dialogues for training and 174 for testing. After splitting, we mean-center the dialogue values in the training set for each attribute and scale them to have unit variance. We apply the same transformation to the test set using the distribution parameters of the training set. 
Our final pre-processing step is to compute the acceptability score. To do this, we compute the covariance matrix of the dialogue-level attribute values in the training set, which include the donation amount and psychological attributes for both the persuader and persuadee from the original dataset, along with our computed sentiment scores. Since the values are all standardized, the covariances are equivalent to Pearson's $r$. We select the covariances of all attributes with respect to the persuadee donation amount (see Figure \ref{covariances} in Appendix \ref{section_training_set_covariances_for_acceptability_score}) and define the acceptability score of each dialogue $D$ as defined in Section \ref{section_methods}. We use the same covariances obtained from the training set to compute acceptability scores on the test set. After pre-processing, the training set has 52 total attributes. These include the persuadee/persuader donation amounts, psychological variables, sentiment, and the acceptability score.

\subsubsection{Progression Model Training} \label{subsubsection_progression_model_training}
We train four progression models as outlined in Sections \ref{subsection_unsupervised_approach} and \ref{subsection_supervised_approach}: (1) Unsupervised, (2) RoBERTa-base, (3) RoBERTa-large, and (4) RoBERTa-large-adapted. For each model, 10\% of the training set is held out as a validation set (58 dialogues). For the unsupervised model, a grid search is run for the hyperparameters (e.g., \# clusters, recency $\beta$, dim. reduction, etc.) over the validation set, and the final model is obtained by re-training over the full training set using the best hyperparameters. The final model uses k-means for clustering with $k=21$ and recency weight $\beta=0.3$. A complete hyperparameter listing and details on the grid search can be found in Appendix \ref{section_grid_search_details_for_unsupervised_model}. For the supervised RoBERTa models, fine-tuning is done with AdamW \citep{loshchilov2018decoupled} and an initial learning rate of $3 \times 10^{-5}$ for a maximum of 30 epochs. Early stopping is used over the validation set with the checkpoint corresponding to the lowest validation loss selected as the final model.

\subsubsection{Automatic Evaluation} \label{subsubsection_automatic_evaluation}
We evaluate the progression models on the following automatic metrics: (1) Mean Absolute Error (\textbf{MAE}) in predicting dialogue acceptability, and (2) Pearson's correlation (\textbf{r}) between overall PF slope and dialogue acceptability. With MAE we validate that the progression function is able to estimate success of a complete dialogue, while PF slope correlation validates that during an ongoing dialogue, progression increases over time for high-success dialogues and decreases over time for low-success dialogues. To measure PF slope correlation, we fit a least-squares regression line to the progression curve of each dialogue in the test set, and measure Pearson's $r$ between the regression slopes and their corresponding acceptability scores. For robustness we repeat this evaluation 33 times with varying initialization seeds for each model type (final hyperparameters stay constant) and report the means and standard deviations across runs in Table \ref{pf_auto_results}.

\begin{table}[t]
    \centering
    \scriptsize
    \caption{Progression Function Auto Eval Results: Mean (SD) of MAE and Pearson's $r$ across runs.}
    \label{pf_auto_results}
    \begin{tabular}{lcc}
        \toprule
        \footnotesize \textbf{Model} & 
        \footnotesize \textbf{MAE} & 
        \footnotesize \textbf{r} \\
        \midrule
        Unsupervised*         &         1.37  ($\pm$ 0.01) &         0.40  ($\pm$ 0.02) \\
        RoBERTa-base          &         0.99  ($\pm$ 0.06) &         0.50  ($\pm$ 0.04) \\
        RoBERTa-large         &         0.97  ($\pm$ 0.10) &         0.54  ($\pm$ 0.06) \\
        RoBERTa-large-adapted & \textbf{0.95} ($\pm$ 0.05) & \textbf{0.57} ($\pm$ 0.03) \\
        \bottomrule
        \multicolumn{3}{l}{For Pearson's $r$, $p < 0.01$ for all runs (two-tailed; $H_0$ is non-correlation).} \\
        \multicolumn{3}{l}{* Hyperparameters of the unsupervised model can be found in Appendix \ref{section_final_unsupervised_model_hyperparameters}.}
    \end{tabular}
    \vskip -0.15in
\end{table}

Unsurprisingly, the supervised models outperform the unsupervised model on both metrics, which is expected since they directly optimize a mean squared error objective. Of the supervised models, the RoBERTa-large instances perform the best, with dialogue domain adaptation boosting each metric.
Pearson's $r$ is significant at the $p < 0.01$ level for all runs (the null hypothesis is non-correlation).

\subsubsection{Manual Evaluation} \label{subsubsection_manual_evaluation}
\begin{figure}[t]
\vskip -0.03in
\begin{center}
\includegraphics[width=\columnwidth]{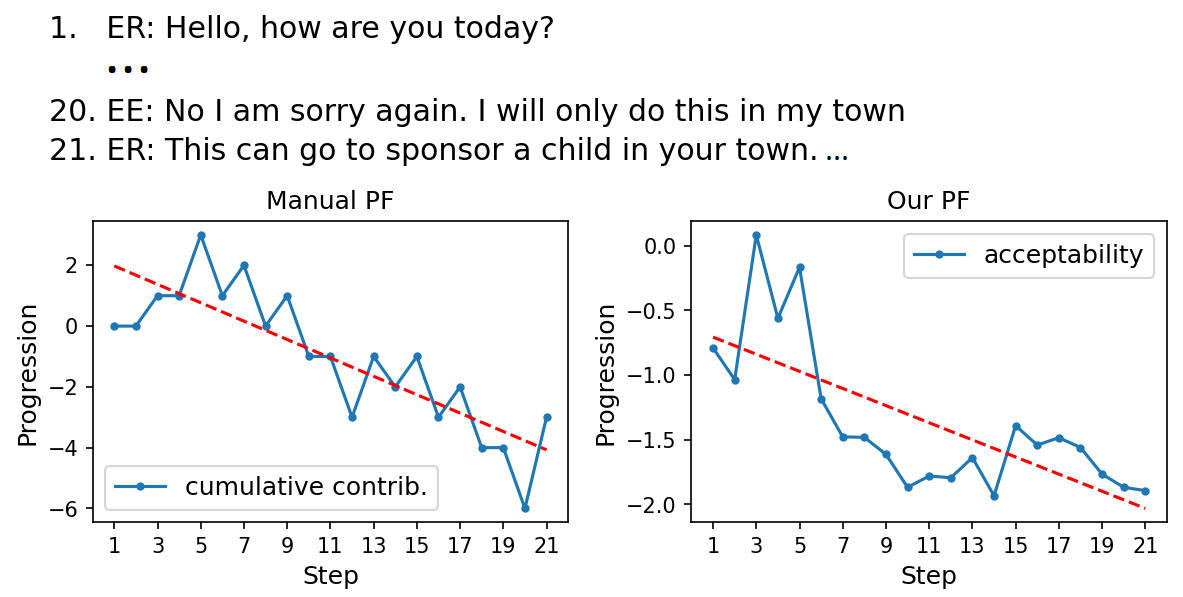}
\caption{Left: ground-truth progression curve given by the cumulative sum of utterance-level manual ratings. Right: estimated curve from RoBERTa-large-adapted.} 
\label{manual_example}
\end{center}
\vskip -0.2in
\end{figure}

To obtain a more precise evaluation, we asked three annotators to estimate sentence-level progression for dialogues in our test set. Two graduate students and one postdoc in our lab served as our annotators. For each of twelve randomly selected test dialogues, each annotator rated all sentences on a scale of \{-1, 0, 1\} for progression, with -1 indicating regression from the task goal, 0 indicating neutral progression, and +1 indicating progression toward the task goal. Altogether, each annotator provided 431 sentence ratings across 244 utterances. After aggregating at the utterance level, average inter-annotator agreement is 0.57 (Cohen’s kappa). For each dialogue, the cumulative sum of the utterance-level manual ratings creates a ground-truth progression curve, as shown in Figure \ref{manual_example}.

We evaluate the progression models against the ground-truth curves using Pearson's correlation ($r$) since the PF output and cumulative manual ratings are continuous and on different scales. We report the following correlations: (1) between utterance-level PF value and ground-truth value (\textbf{utt}), (2) between utterance-level PF slope and ground-truth slope (\textbf{utt-sl}), (3) between dialogue-level PF slope and ground-truth slope (\textbf{dlg-sl}), and (4) between dialogue-level PF slope and the final ground-truth progression value (\textbf{dlg-sl-f}). We repeat this evaluation for each of the 33 uniquely initialized model instances of each type from Section \ref{subsubsection_automatic_evaluation}, averaging each metric across raters and then across runs. We report the aggregate means and standard deviations across runs in Table \ref{pf_manual_results}. A complete listing of results for each rater (averaged across runs) is provided in Appendix \ref{section_full_manual_evaluation_results}, along with detailed explanations of each metric (Appendix \ref{section_explanations_of_manual_metrics}). Additionally, Figure \ref{manual_example_pf_is_right} in Appendix \ref{section_pf_disagreement_examples} provides examples of disagreement between the PF and ground-truth progression curves which can be challenged despite high inter-annotator agreement, demonstrating difficulty in establishing ground-truth for this open-ended task.

\begin{table}[t]
    \centering
    \scriptsize
    \caption{Progression Function Manual Eval Results: Mean (SD) of \textbf{utt}, \textbf{utt-sl}, \textbf{dlg-sl}, \& \textbf{dlg-sl-f} across runs.}
    \label{pf_manual_results}
    \begin{tabular}{l@{\hspace{0.5\tabcolsep}}c@{\hspace{0.5\tabcolsep}}c@{\hspace{0.5\tabcolsep}}c@{\hspace{0.5\tabcolsep}}c}
        \toprule
        \footnotesize \textbf{Model} & 
        \footnotesize \textbf{utt} & 
        \footnotesize \textbf{utt-sl} & 
        \footnotesize \textbf{dlg-sl} & 
        \footnotesize \textbf{dlg-sl-f} \\
        \midrule
        Unsuperv.   &         0.09  ($\pm$ 0.13) &         0.04  ($\pm$ 0.01) &        -0.01  ($\pm$ 0.07) &        -0.07  ($\pm$ 0.07) \\
        RoBERTa-b   &         0.39  ($\pm$ 0.07) &         0.18  ($\pm$ 0.03) &         0.26  ($\pm$ 0.06) &         0.26  ($\pm$ 0.06) \\
        RoBERTa-l   &         0.39  ($\pm$ 0.08) &         0.18  ($\pm$ 0.06) &         0.34  ($\pm$ 0.09) &         0.35  ($\pm$ 0.10) \\
        RoBERTa-l-a & \textbf{0.48} ($\pm$ 0.05) & \textbf{0.24} ($\pm$ 0.03) & \textbf{0.43} ($\pm$ 0.06) & \textbf{0.41} ($\pm$ 0.07) \\
        \bottomrule
        \multicolumn{5}{l}{b = base; l = large; l-a = large-adapted.}
    \end{tabular}
\end{table}

\subsubsection{Benefit of Domain Adaptation}
To verify the beneficial effect of domain adaptation we perform two-tailed paired t-tests to confirm the differences in means between RoBERTa-large and RoBERTa-large-adapted on all automatic and manual metrics. For each metric, we pair the results from both models for each run of the same seed, since their regression heads would have received identical initializations. We find that the means of \textbf{utt}, \textbf{utt-sl}, \textbf{dlg-sl}, and \textbf{dlg-sl-f} differ at the $p < 0.01$ significance level, and the means of the automatic Pearson's $r$ metric differ at the $p < 0.05$ level. This confirms our intuition that domain adaptation for dialogue prior to fine-tuning the regression objective aids generalization in this task.

\subsection{Rollout Experiments} \label{subsection_rollout_experiments}
To demonstrate the ability of the PF to guide a dialogue agent, we use it to score rollouts generated with DialoGPT as described in Section \ref{subsection_response_planning}. Specifically, we design a self-play experiment to automatically evaluate the effect of PF-guided rollouts on the success of the solicitation task in Persuasion For Good. The following summarizes  the experimental setup, procedure, and results.

\subsubsection{Exeperimental Setup}\label{subsubsection_experimental_setup}
First, we fine-tune DialoGPT to generate responses on Persuasion For Good. We add speaker control tokens to the vocabulary so that the model can be conditioned to generate as the persuader or persuadee, respectively. Training is done with AdamW (initial lr=$5\times 10^{-5}$) for 6 epochs with early stopping over a 10\% validation set using perplexity.
The final model checkpoint was selected after 3 epochs, achieving validation perplexity of 8.82.

We then select a progression model to use for all self-play runs. Since the supervised RoBERTa-large-adapted model achieved the best average scores across all automatic and manual evaluations, we randomly select one of the 33 RoBERTa-large-adapted instances from Section \ref{subsubsection_automatic_evaluation} to use for all runs. We use this instance for rollout scoring and to measure the progression of each self-play dialogue.

Finally, we train a binary classifier to identify if the persuadee has stated the intent to donate in a conversation, which we use to detect successful self-play dialogues. We fine-tune a RoBERTa model as a classifier using just the persuadee's utterances as input and use binarized donation labels in Persuasion For Good as targets. Specifically, for each dialogue the label is 0 if the donated amount is \$0, otherwise it is 1. We use the manually verified intended donation labels from Persuasion For Good ``ANNSET'' for our validation and test splits and use the remaining end-task donation labels for training. Training is done with early stopping over the validation split using macro F1. The final model checkpoint achieved test F1 of 0.89 and test accuracy of 0.90.
All three trained models used in this experiment are available to download (see our code release for instructions and hyperparameters).

\subsubsection{Self-Play Procedure} \label{subsubsection_self_play_procedure}
From our test set of 174 dialogues, we manually filter out 41 in which the persuadee pledges a donation within the first 10 utterances, leaving 133 remaining conversations. For each of these, the response generator is given the first 10 utterances as context and tasked to complete a second set of 10 utterances, playing the role of both the persuader and persuadee. Since the task is solicitation, we allow the generator to use rollouts only when acting as the persuader. We perform the self-play experiment using three persuader modes: (1) with no rollouts (\textbf{No RO}), (2) with 2 response candidates, 2 rollouts per candidate, and 3 utterances per rollout (\textbf{2x2x3}), and (3) with 3 response candidates, 3 rollouts per candidate, and 5 utterances per rollout (\textbf{3x3x5}). For each utterance in each rollout, we use beam sampling with num\_beams=6, top\_k=50, top\_p=0.95, and temperature=$1.5+0.002\cdot T$ where $T$ is the number of tokens in the dialogue history. After generation, we compute the following metrics for each dialogue: (1) progression using the selected RoBERTa-large-adapted instance (\textbf{Prog.}), (2) persuader and persuadee sentiment using the sentiment classifier from Section \ref{subsection_progression_function_experiments} (\textbf{ER Sent.} \& \textbf{EE Sent.}), and (3) the percent of test dialogues where the persuadee pledges a donation amount or explicitly states intent to donate, as detected by the binary donation intent classifier (\textbf{EE Don.\%}). 

\subsubsection{Self-Play Results}\label{subsubsection_self_play_results}

\begin{table}[t]
    \centering
    \scriptsize
    \caption{Rollouts self-play results: Mean (SD) of progression, sentiment, and \% donated across runs. }
    \label{rollouts_self_play_results}
    \begin{tabular}{l@{\hspace{1.25\tabcolsep}}c@{\hspace{1.25\tabcolsep}}c@{\hspace{1.25\tabcolsep}}c@{\hspace{1.25\tabcolsep}}l}
        \toprule
        \footnotesize \textbf{Mode} & 
        \footnotesize \textbf{Prog.} & 
        \footnotesize \textbf{ER Sent.} & 
        \footnotesize \textbf{EE Sent.} & 
        \footnotesize \textbf{EE Don. \%} \\
        \midrule
        No RO  &         0.01  ($\pm$ 0.24) &         0.51  ($\pm$ 0.03) &         0.44  ($\pm$ 0.06) &         38\%  ($\pm$ 6\%) \\
        \midrule
        2x2x3  &         0.69  ($\pm$ 0.29) &         0.62  ($\pm$ 0.05) &         0.49  ($\pm$ 0.07) &         45\%  ($\pm$ 10\%)\dag \\
        3x3x5  & \textbf{0.95} ($\pm$ 0.16) & \textbf{0.66} ($\pm$ 0.02) & \textbf{0.52} ($\pm$ 0.04) & \textbf{47\%} ($\pm$ 11\%) \\
        \bottomrule
        \multicolumn{5}{l}{All 2x2x3 and 3x3x5 means significant at $p < 0.01$ (or 0.05 if marked \dag) }\\
        \multicolumn{5}{l}{when compared to No RO with two-tailed paired t-tests. ER and EE refer}\\
        \multicolumn{5}{l}{  to the persuader and persuadee respectively.}
    \end{tabular}
    \vskip -0.15in
\end{table}

For robustness we repeat this procedure 5 times with varying generation seeds for each persuader mode. In total, 1,995 self-play dialogues are completed (133 dialogues for each of 3 modes for each of 5 seeds). We average each metric across dialogues and then across runs, and report the aggregate means and standard deviations across runs.
Additionally, to verify the benefit of rollouts, we perform two-tailed paired t-tests to confirm the differences in means between the rollout-enabled modes (\textbf{2x2x3} and \textbf{3x3x5}) and the baseline (\textbf{No RO}). For each metric, we average the results across runs and pair these averages from both modes for each dialogue. Results are shown in Table \ref{rollouts_self_play_results}.

We observe that the mean progression increases significantly when rollouts are used. This is expected since response candidates with the highest average end-rollout progression are selected. We also observe that rollouts lead to higher average sentiment for both the persuader and persuadee, which makes sense due to the correlation between sentiment and the acceptability score (see Figure \ref{covariances} in Appendix \ref{section_training_set_covariances_for_acceptability_score}). Finally, rollouts yield a higher percentage of dialogues with a pledged or intended donation. \footnote{We randomly selected 30 of the 1,995 self-play dialogues and manually inspected them for donation intent. We found the classifier labeled 26 of them correctly (87\% accuracy).} All of these results are significant at the $p < 0.01$ level except for \textbf{EE Don.\%} in \textbf{2x2x3} mode which is significant at $p < 0.05$.

Although progression is noticeably higher for the \textbf{3x3x5} mode than for the \textbf{2x2x3} mode (0.95 vs 0.69), all other metrics are close between these modes with a small advantage in \textbf{3x3x5} mode.  This suggests that scaling rollout simulations can be beneficial, but there may be diminishing returns for simulation size. 
Example self-play dialogues are provided in Tables \ref{response_generation_example_1}, \ref{response_generation_example_2}, and \ref{response_generation_example_3} in Appendix \ref{section_rollout_self_play_examples}.

\section{Limitations \& Future Direction} \label{section_limitations_future_directions}
We recognize several limitations of our study which warrant follow-up investigation. This study focuses on a single task and dataset, and thus is subject to the assumptions and biases therein. Since we intend our framework to be general, it is prudent to perform additional studies to verify the efficacy of our methods on a variety of datasets spanning multiple dialogue domains and tasks. Also, although we provide automatic evaluation of the ability of rollouts to improve performance on a solicitation task, we cannot assume that humans would respond in the same way as DialoGPT. Thus, human evaluation is needed to further validate this approach.

\section{Conclusion} \label{section_conclusion}
In this work we introduced the concept of global dialogue state and proposed a framework with which a dialogue agent can gain awareness of where an ongoing conversation is headed, the likelihood of a successful outcome, and how its own response decisions impact the overall direction of the dialogue. We demonstrated that an unsupervised approach to modeling the GDS space and progression function is feasible, which is useful in sparsely-labeled settings. However, we showed that with domain-adaptation pre-training for dialogue, supervised methods are preferable when labels are available. Finally, we demonstrated how using the PF as a feedback mechanism via dialogue rollouts allows an agent to improve outcomes on a solicitation task.

\section*{Ethical Considerations}
\subsection*{Ethical Dialogue Systems}
We acknowledge the potential risks inherent in the deployment of goal-oriented dialogue systems, and especially note that care must be taken to ensure persuasive dialogue systems are designed for beneficial use as discussed by \citet{Wan:19}. Concretely, when applying our framework, care must be taken to ensure that the goal of the system (defined by the primary success attribute of the acceptability score) should be generally accepted as beneficial. For example, our basis for dialogue acceptability in this work is with respect to raising money for children's charity. In general, the achievement of the system's goal should not intentionally lead the user or any other party to harm. Additionally, the definition of acceptability, through its primary or any other correlated attributes, should not allow for discriminative responses, purposefully malicious discourse, or other violations of accepted ethical standards. For example, we include sentiment as secondary attributes in the acceptability score, which, when applied via dialogue rollouts, encourages the system to be courteous, polite, and respectful. It is possible with minimal effort to include further secondary attributes that identify bias, hate speech, and other indicators to help the system remain safe to use.

\subsection*{Annotator Compensation}
All manual annotators were recruited on a voluntary basis in an educational setting and did not receive or expect monetary compensation. Specifically, two graduate students and one postdoc in our lab served as our annotators.

\subsection*{Environmental Impact}
All training and inference in this work was done with two NVIDIA Quadro RTX 8000 GPUs. The most compute-intensive portion of the work was the additional domain adaptation pre-training for RoBERTa-large-adapted (see Section \ref{subsection_supervised_approach}), which took approximately two weeks. After that the multi-seed self-play evaluations took approximately four days, and all other operations (e.g., training and evaluating PF models, fine-tuning DialoGPT) took 24 hours or less.

\section*{Acknowledgements}
We would like to thank our manual annotators for their valuable contribution and the anonymous reviewers for their helpful feedback.
This paper is based upon work supported in part by the United States Air Force under Contract No. FA8750-21-C-0075 and in part by the IBM Corporation under the Artificial Intelligence Research Collaboration Agreement No. W1771793 between IBM and Rensselaer. Any opinions, findings and conclusions or recommendations expressed in this material are those of the author(s) and do not necessarily reflect the views of USAF or IBM Corporation.

\bibliography{anthology,custom}
\bibliographystyle{acl_natbib}

\newpage
\onecolumn

\appendix

\section{Software Packages Used}
\label{section_software_packages_used}
\begin{table}[h]
    \centering
    \scriptsize
    \caption{Software packages used in obtaining or presenting the results in this work}
    \label{software_packages}
    \begin{tabular}{llll}
        \toprule
        \textbf{Package} & \textbf{Version} & \textbf{Citation} & \textbf{URL} \\ 
        \midrule
        hdbscan & 0.8.27 & \citep{mcinnes2017hdbscan} & \url{https://hdbscan.readthedocs.io/} \\
        kneed & 0.7.0 & \citep{5961514} & \url{https://kneed.readthedocs.io/} \\
        Matplotlib & 3.3.4 & \citep{Hunter:2007} & \url{https://matplotlib.org/} \\
        NumPy & 1.19.5 & \citep{harris2020array} & \url{https://numpy.org/} \\
        Pandas & 1.2.4 & \citep{reback2020pandas} & \url{https://pandas.pydata.org/} \\
        plotly & 5.1.0 & \citep{plotly} & \url{https://plotly.com/python/} \\
        PyTorch & 1.9.0 & \citep{NEURIPS2019_9015} & \url{https://pytorch.org/} \\
        PyTorch Lightning & 1.5.8 & \citep{pytorchlightning} & \url{https://pytorchlightning.ai} \\
        scikit-learn & 0.24.1 & \citep{scikit-learn} & \url{https://scikit-learn.org/} \\
        SciPy & 1.6.2 & \citep{2020SciPy-NMeth} & \url{https://scipy.org/scipylib/index.html} \\
        Sentence-Transformers & N/A* & \citep{reimers-2019-sentence-bert} & \url{https://sbert.net/} \\
        TensorFlow & 2.5.1 & \citep{tensorflow2015-whitepaper} & \url{https://tensorflow.org/} \\
        Transformers & 4.11.3 & \citep{wolf-etal-2020-transformers} & \url{https://huggingface.co/transformers/} \\
        umap-learn & 0.5.1 & \citep{mcinnes2018umap-software} & \url{https://umap-learn.readthedocs.io/} \\
        \bottomrule
        \multicolumn{4}{l}{* we use all-mpnet-base-v2 directly through Transformers, but it is part of the Sentence-Transformers model library. Additionally, we base}\\
        \multicolumn{4}{l}{\ \ \ parts of our sentence embedding implementation on that found in Sentence-Transformers.}
    \end{tabular}
\end{table}

\section{Training Set Covariances For Acceptability Score}\label{section_training_set_covariances_for_acceptability_score}
\begin{figure}[h]
\begin{center}
\includegraphics[width=\textwidth]{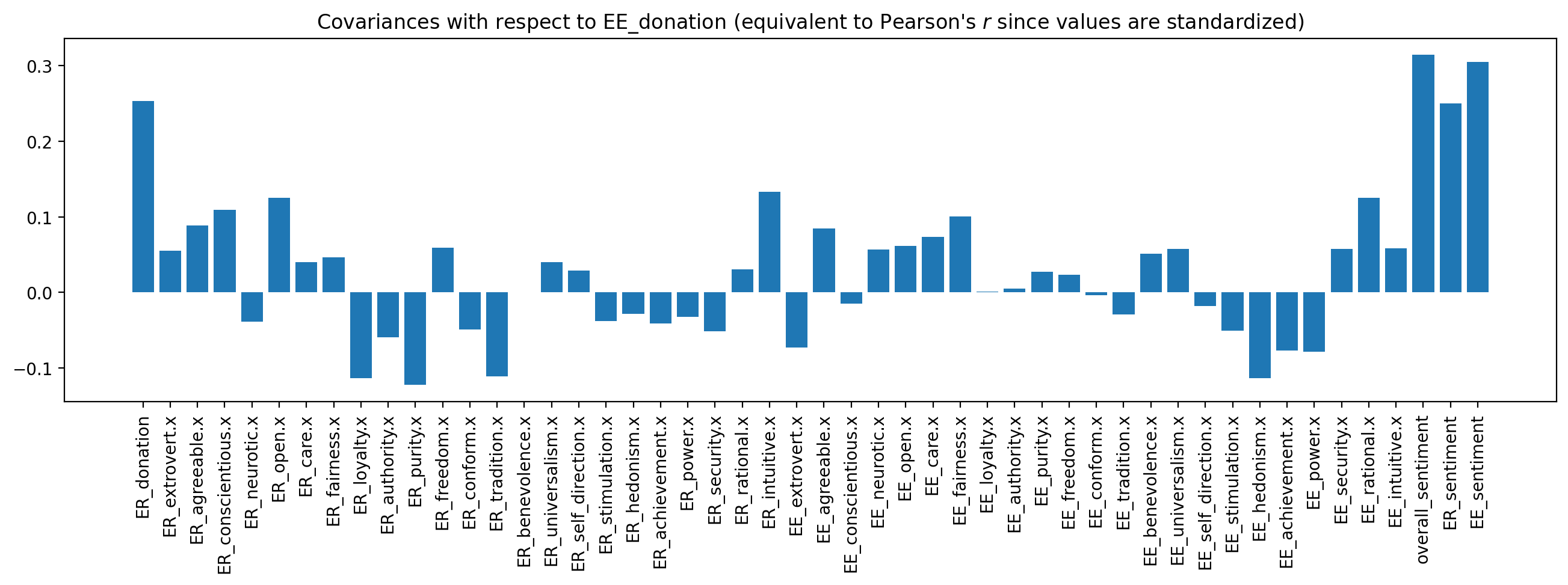}
\caption{The covariances of all other dialogue attributes with respect to the persuadee donation are used to weight the acceptability score. ER and EE refer to the persuader and persuadee respectively.}
\label{covariances}
\end{center}
\vskip -0.2in
\end{figure} 

\section{Full Manual Evaluation Results}
\label{section_full_manual_evaluation_results}

\begin{table}[h]
    \centering
    \footnotesize
    \caption{Progression Function Manual Eval Results (all annotators, averaged across all runs)}
    \label{pf_manual_results_all_annotators}
    \begin{tabular}{lcccc}
        \toprule
        \textbf{Model} & \textbf{utt (1/2/3)} & \textbf{utt-sl (1/2/3)} & \textbf{dlg-sl (1/2/3)} & \textbf{dlg-sl-f (1/2/3)} \\
        \midrule
        unsupervised          & 0.09      / 0.06      / 0.12        & 0.02      / 0.04      / 0.05      & 0.04 / -0.03    / -0.02    & -0.07 / -0.09 / -0.05 \\
        RoBERTa-base          & 0.39\ddag / 0.30\ddag / 0.48\ddag   & 0.15\dag  / 0.17\dag  / 0.21\ddag & 0.30 / 0.12     / 0.34     & 0.26  / 0.16  / 0.35  \\
        RoBERTa-large         & 0.39\ddag / 0.30\ddag / 0.50\ddag   & 0.16      / 0.18\dag  / 0.21\dag  & 0.41 / 0.17     / 0.46     & 0.36  / 0.21  / 0.47  \\
        RoBERTa-large-adapted & \textbf{0.49}\ddag / \textbf{0.37}\ddag / \textbf{0.59}\ddag   & \textbf{0.21}\ddag / \textbf{0.24}\ddag / \textbf{0.29}\ddag & \textbf{0.51} / \textbf{0.26}     / \textbf{0.52}     & \textbf{0.45}  / \textbf{0.27}  / \textbf{0.51}  \\
        \bottomrule
        \multicolumn{5}{l}{\scriptsize Average Pearson's $r$ p-value across runs: \dag: $p < 0.05$;\ \ \ddag: $p < 0.01$; (two-tailed; $H_0$ is non-correlation). }
    \end{tabular}
\end{table}

\section{Explanations of Manual Metrics}\label{section_explanations_of_manual_metrics}

\begin{figure}[h]
\vskip -0.25in
\begin{center}
\includegraphics[width=\textwidth]{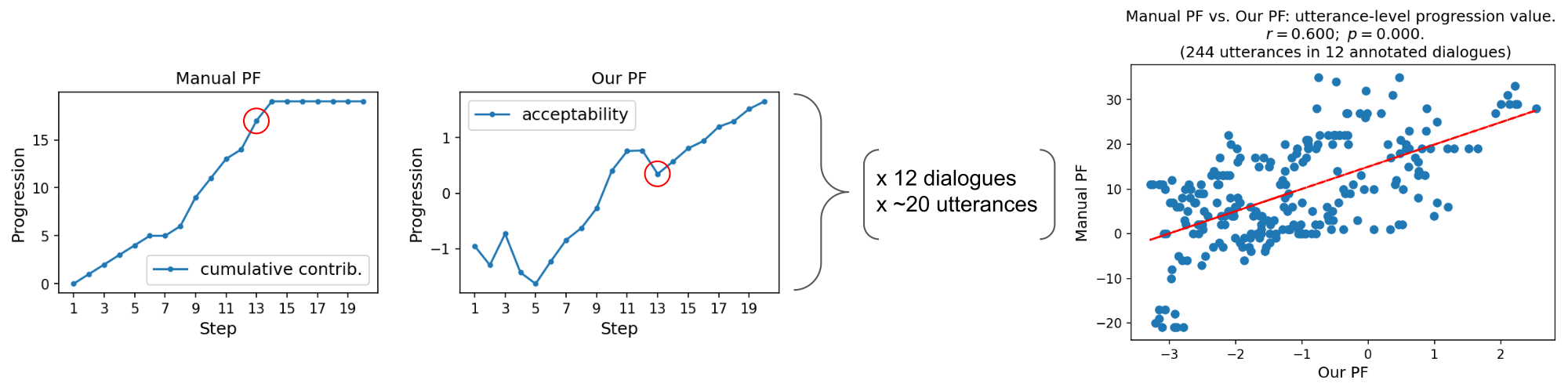}
\caption{\textbf{utt}: Pearson's $r$ (right) between utterance-level PF values (center, e.g., circled) and ground-truth values (left, e.g., circled) for all 244 utterances across 12 dialogues. Points shown on the right are from annotator 3. This metric is intended to measure if the PF and ground-truth progression curves assign similar values (relative to their respective scales) at each step of an ongoing dialogue.}
\label{manual_eval_utt_r}
\end{center}
\vskip -0.2in
\end{figure} 

\begin{figure}[h]
\vskip -0.25in
\begin{center}
\includegraphics[width=\textwidth]{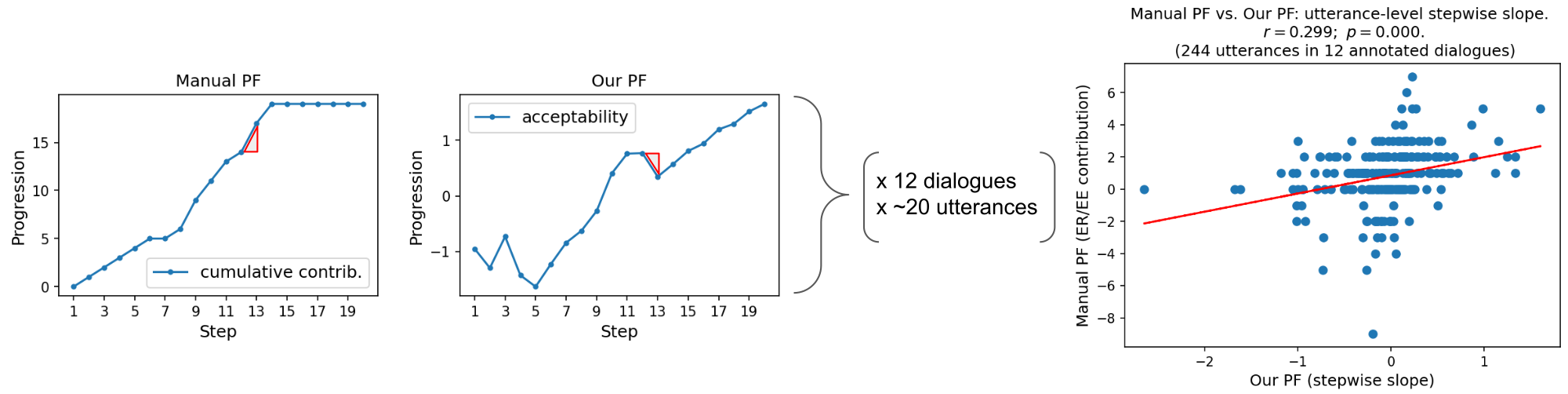}
\caption{\textbf{utt-sl}: Pearson's $r$ (right) between utterance-level PF slopes (center, e.g., see triangle) and ground-truth slopes (left, e.g., see triangle), for all 244 utterances across 12 dialogues. Utterance-level slopes are computed as the differences in the progression curves between two dialogue steps.  Points shown on the right are from annotator 3. This metric is intended to measure if the PF and ground-truth progression curves move in the same direction at each step of an ongoing dialogue.}
\label{manual_eval_utt_slope_r}
\end{center}
\vskip -0.2in
\end{figure} 

\begin{figure}[h]
\vskip -0.25in
\begin{center}
\includegraphics[width=\textwidth]{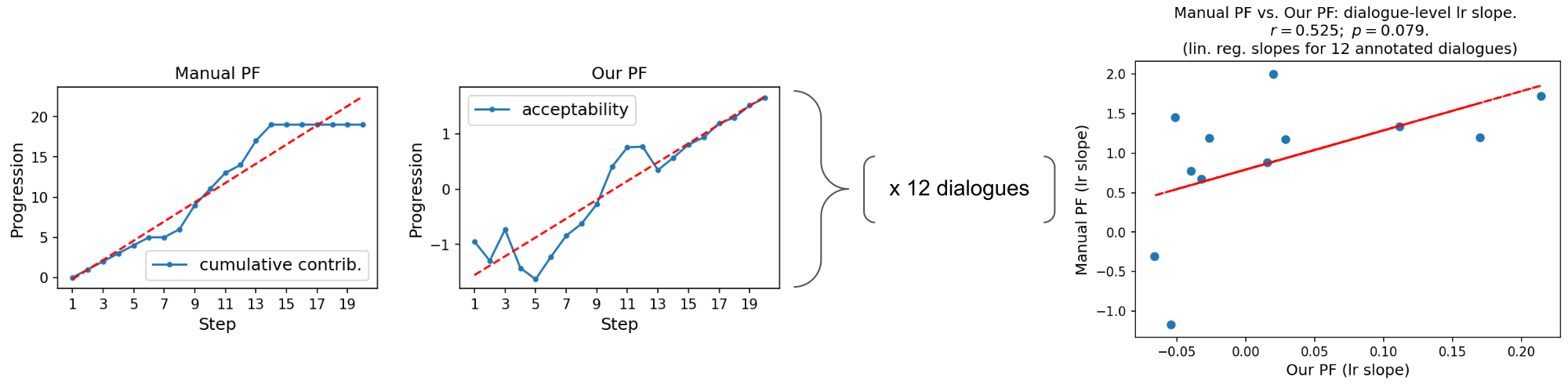}
\caption{\textbf{dlg-sl}: Pearson's $r$ (right) between dialogue-level PF slopes (center, e.g., see line) and ground-truth slopes (left, e.g., see line), for all 12 dialogues. Dialogue-level slopes are computed by fitting least-squares regression lines to the progression curves.  Points shown on the right are from annotator 3. This metric is intended to measure the ability of the overall PF trend to approximate the ground-truth progression curve.}
\label{manual_eval_dlg_slope_r}
\end{center}
\vskip -0.2in
\end{figure}

\begin{figure}[h]
\vskip -0.25in
\begin{center}
\includegraphics[width=\textwidth]{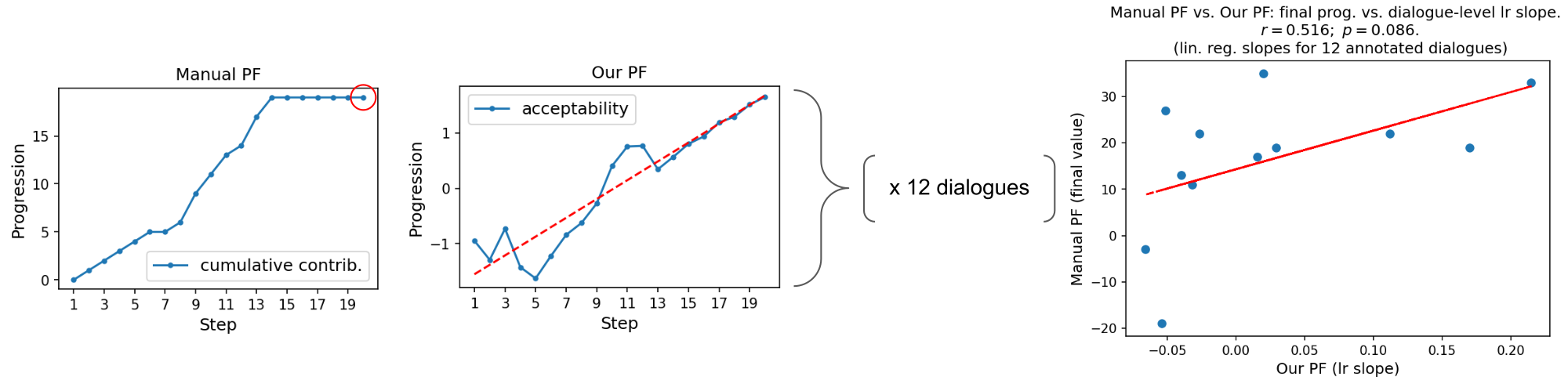}
\caption{\textbf{dlg-sl-f}: Pearson's $r$ (right) between dialogue-level PF slopes (center, e.g., see line) and the final ground-truth progression value (left, e.g., circled), for all 12 dialogues. Dialogue-level slopes are computed by fitting least-squares regression lines to the progression curves.  Points shown on the right are from annotator 3. This metric is intended to measure the ability of the overall PF trend to predict the end-task outcome.}
\label{manual_eval_dlg_slope_final_r}
\end{center}
\vskip -0.2in
\end{figure}

\clearpage

\section{PF Disagreement Examples}\label{section_pf_disagreement_examples}

\begin{figure}[h]
\vskip 0.05in
\begin{center}
\includegraphics[width=\textwidth]{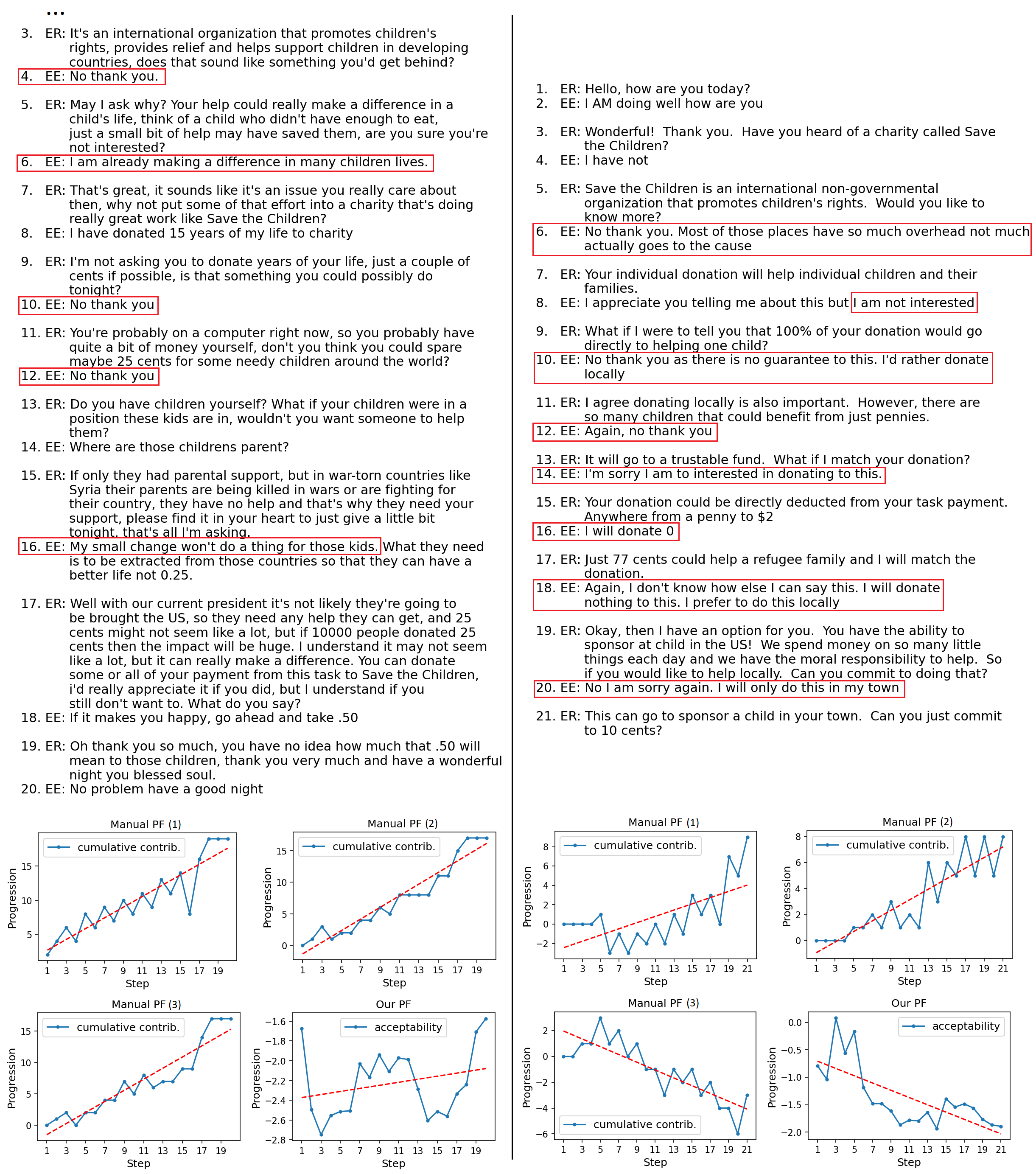}
\caption{Two examples where the PF (RoBERTa-large-adapted) disagrees with multiple annotators. We note that the PF correctly decreases (\textbf{Right}) and stays negative (\textbf{Left}) given the turns boxed in red showing poor progression. }
\label{manual_example_pf_is_right}
\end{center}
\vskip -0.2in
\end{figure}

\clearpage

\section{Grid Search Details for Unsupervised Model}\label{section_grid_search_details_for_unsupervised_model}

\begin{algorithm}[h]
    \caption{Grid search for hyperparameter tuning of the unsupervised progression model on the validation set. Descriptions for each hyperparameter are provided in Table \ref{hyperparam_descriptions}.}
    \SetAlgoLined
    \For{$\beta \in \{0.0, 0.1, \dots, 2.0 \}$} { 
        \For{$d \in \{2, 16, 32, 64, 128, 768 \}$} {
            \For{$\text{normalize\_embeddings} \in \{\text{True}, \text{False}\}$} {
                \For{$\text{distance\_metric} \in \{\text{Cosine}, \text{Euclidean}\}$} {
                    \Comment{k-means experiments}
                    \For{$k \in \{2, 3, \dots, 30 \}$} {
                        \For{$\text{inverse\_distance} \in \{\text{True}, \text{False}\}$} {
                            \For{$\text{standardized\_proximity} \in \{\text{True}, \text{False}\}$} {
                                measure\_PF\_slope\_r()\;
                            }
                        }
                    }
                    \Comment{HDBSCAN experiments}
                    \For{$\text{min\_cluster\_size} \in \{10, 20, \dots, 100 \}$} {
                        \For{$\text{soft\_value\_aggregation} \in \{\text{True}, \text{False}\}$} {
                            \For{$\text{prob\_scaling} \in \{\text{None}, \text{softmax}, \text{sum}\}$} {
                                \For{$\text{standardized\_proximity} \in \{\text{True}, \text{False}\}$} {
                                    measure\_PF\_slope\_r()\;
                                }
                            }
                        }
                    }
                }
            }
        }
    }
\end{algorithm}

\begin{table}[h]
    \centering
    \scriptsize
    \caption{Hyperparameter Descriptions}
    \label{hyperparam_descriptions}
    \begin{tabular}{ll}
        \toprule
        \textbf{Hyperparameter} & \textbf{Description} \\ 
        \midrule
        $\beta$ (recency weight) & Controls how much emphasis is placed on recent tokens when computing dialogue embeddings.\\
        $d$ (embedding size) & The dimensionality of dialogue embeddings. Values $< 768$ reduced with Parametric UMAP. \\
        normalize\_embeddings & If True, embeddings are normalized to have unit magnitude after dimensionality reduction.\\
        distance\_metric & The distance metric used by Parametric UMAP and centroid proximity calculations.* \\
        k (number of clusters) & Number of clusters to use for k-means.\\
        inverse\_distance & If True, Euclidean centroid proximity is computed as the inverse distance instead of negative distance.** \\
        standardized\_proximity & If True, centroid proximities are converted to z-scores before progression is computed. \dag\\
        min\_cluster\_size & Minimum number of points in a HDBSCAN cluster. Clusters with fewer points get merged into larger ones. \\
        soft\_value\_aggregation & If True, HDBSCAN cluster attribute aggregations are weighted with cluster membership probabilities.\\
        prob\_scaling & The type of scaling applied to progression computed from HDBSCAN cluster membership probabilities. \\
        \bottomrule
        \multicolumn{2}{l}{*: Centroid proximity calculations refer to progression computation. Proximities used during clustering are always Euclidean.}\\
        \multicolumn{2}{l}{**: Does not apply to cosine distance.}\\
        \multicolumn{2}{l}{\dag: Applies only if k-means is used, or if HDBSCAN is used with softmax probability scaling.}
    \end{tabular}
\end{table}

\bigbreak
\bigbreak

\section{Final Unsupervised Model Hyperparameters}\label{section_final_unsupervised_model_hyperparameters}
The final unsupervised model uses k-means ($k=21$), $\beta=0.3$, $d=768$, normalized embeddings, euclidean distance, and inverse distance for centroid proximity.

\bigbreak
\bigbreak

\section{Examples from rollout self-play experiment}\label{section_rollout_self_play_examples}
The following examples in Tables \ref{response_generation_example_1}, \ref{response_generation_example_2}, and \ref{response_generation_example_3} compare self-play between the baseline \textbf{No RO} mode and the two rollout modes \textbf{2x2x3} and \textbf{3x3x5}. In each example, the utterance where the persuadee (EE) states intent to donate is highlighted in \textcolor{red}{red}. The complete dialogues from the self-play experiments are included with our code release.

\begin{table*}[t]
    \vskip -0.20in
    \scriptsize
    \caption{Rollout Self-Play Example 1: \textbf{No RO} vs. \textbf{2x2x3} (dialogue \#7; generation seed: 247)}
    \label{response_generation_example_1}
    \begin{tabular}{l}
    \toprule
    \textbf{CONTEXT:} \ \ \ \ \ \ \ \ \ \ \ \ \ \ \ \ \ \ \ \ \ \ \ \ \ \ \ \ \ \ \ \ \ \ \ \ \ \ \ \ \ \ \ \ \ \ \ \ \ \ \ \ \ \ \ \ \ \ \ \ \ \ \ \ \ \ \ \ \ \ \ \ \ \ \ \ \ \ \ \ \ \ \ \ \ \ \ \ \ \ \ \ \ \ \ \ \ \ \ \ \ \ \ \ \ \ \ \ \ \ \ \ \ \ \ \ \ \ \ \ \ \ \ \ \ \ \ \ \ \ \ \ \ \ \ \ \ \ \ \ \ \ \ \ \ \ \ \ \ \ \ \ \ \ \ \ \ \ \ \ \ \ \ \ \ \ \ \ \ \ \ \ \ \ \ \ \ \ \ \ \ \ \ \ \ \ \ \ \ \ \ \ \ \ \ \ \ \ \ \ \ \ \ \ \ \ \ \ \ \ \ \ \ \ \ \ \ \ \ \ \ \ \ \ \\
    \midrule
    ER: Hello, Have you ever donated to a charity? \\
    EE: A few times, yes.\\
    ER: ow was your experience?\\
    EE: It was ok. I didn't really think much of it. Most of the time its just a dollar or two.\\
    ER: Do you donate all te time?\\
    EE: Not so much.\\
    ER: Have yu heard of Save the children? It is pretty well known around the world.\\
    EE: I think so.\\
    ER: What do you think that they do?\\
    EE: Give money to child. Support efforts in war torn countries. Healthcare. All that stuff.\\
    \midrule
    \textbf{NO ROLLOUTS (progression: -2.65; ER sentiment: 0.69; EE sentiment: 0.03)}\\
    \midrule
    ER: I.  It is a worthy cause.  You can donate all you want from it.\\
    EE: I'm sorry. I'm broke right now.  That's why I'm asking if you do donate all the time.\\
    ER: It's okay.  I understand and agree.  I think it's a good cause.\\
    EE: So what is the charity that you donate to?\\
    ER: I don't have a specific charity, but I do think they are a decent charity.\\
    EE: Are they a non-profit?  What do they do?\\
    ER: It's a non profit.  They help protect childrens rights to health and education.\\
    EE: Well, I think they are a great charity.  Do they provide help in other ways besides donating to them?\\
    ER: Yes they help with children in the US and other developing countries.\\
    EE: So I'm guessing the money goes towards programs to help those children get a proper education?\\
    \midrule
    \textbf{2x2x3 ROLLOUTS (progression: 0.15; ER sentiment: 0.67; EE sentiment: 0.63)}\\
    \midrule
    ER: I, they do. I have donated to the Red Cross and the Salvation Army in the past.\\
    EE: I donated a couple of months ago. Do you donate often? I donate when I can.\\
    ER: Yes, I donate to them as well. How much do you like to donate?\\
    EE: i donate \$.50 each month. Do you do the same?\\
    ER: Yeah, that sounds great. Any bit helps though. Do you have an amount in mind?\\
    EE: \textcolor{red}{I would like to donate half. Is that a good idea?}\\
    ER: Yes that would be great. The children really do need it. I donate to them as well.\\
    EE: I'm glad we can agree on a generous donation. How much would you like me to donate?\\
    ER: I think half would be fine, I'm not sure, but that is a very generous amount.\\
    EE: Well, it is nice to think that I'm not getting hit by a bot, so that's a good thing.\\
    \bottomrule
    \end{tabular}
\end{table*}

\begin{table*}[t]
    \vskip -0.20in
    \scriptsize
    \caption{Rollout Self-Play Example 2: \textbf{No RO} vs. \textbf{3x3x5} (dialogue \#88; generation seed: 756)}
    \label{response_generation_example_2}
    \begin{tabular}{l}
    \toprule
    \textbf{CONTEXT:} \ \ \ \ \ \ \ \ \ \ \ \ \ \ \ \ \ \ \ \ \ \ \ \ \ \ \ \ \ \ \ \ \ \ \ \ \ \ \ \ \ \ \ \ \ \ \ \ \ \ \ \ \ \ \ \ \ \ \ \ \ \ \ \ \ \ \ \ \ \ \ \ \ \ \ \ \ \ \ \ \ \ \ \ \ \ \ \ \ \ \ \ \ \ \ \ \ \ \ \ \ \ \ \ \ \ \ \ \ \ \ \ \ \ \ \ \ \ \ \ \ \ \ \ \ \ \ \ \ \ \ \ \ \ \ \ \ \ \ \ \ \ \ \ \ \ \ \ \ \ \ \ \ \ \ \ \ \ \ \ \ \ \ \ \ \ \ \ \ \ \ \ \ \ \ \ \ \ \ \ \ \ \ \ \ \ \ \ \ \ \ \ \ \ \ \ \ \ \ \ \ \ \ \ \ \ \ \ \ \ \ \ \ \ \ \ \ \ \ \ \ \ \ \ \\
    \midrule
    ER: Hello therehow are you?\\
    EE: I am good. How are you?\\
    ER: I'm pretty good, just trying to get ready for work soon.\\
    EE: I understand that.  So tell me about this charity.\\
    ER: Have you heard of Save the Children?  They do work around health, education, etc.\\
    EE: I'm not sure that I have.\\
    ER: They work mostly in developing countries.  I would appreciate it if some of the funds could be donated.  The researchers will pass that donation \\
    \ \ \ \ \ \ \ on to the charity at the end.\\
    EE: What kind of work do they do?\\
    ER: They do emergency hunger relief, are engaged with Syrian children.\\
    EE: Ok, that sounds good.  How do I know it is a legitimate charity?\\
    \midrule
    \textbf{NO ROLLOUTS (progression: -0.43; ER sentiment: 0.71; EE sentiment: 0.50)}\\
    \midrule
    ER: They are a good non-profit that helps children all around the world\\
    EE: Ah, that's good to know.  Do you know what the funding for this charity actually goes towards?\\
    ER: They actually make a huge impact.  Do you donate to charity often?\\
    EE: No, I don't think so, but I will look them up.\\
    ER: They are a 501c, so it's not like they're going to ask you to give anything at this time.  Every little bit helps.\\
    EE: Cool, I don't think I'm ready to give anything up yet.  How about you?\\
    ER: They are a legit charity that I donate to regularly.  Have you donated to a charity before?\\
    EE: No, I'm usually a pretty cautious individual with donations, but I think that would be a good idea.\\
    ER: They're great with giving to charities but I wouldn't be comfortable donating more than the 10.00, but I usually donate my entire earnings \\
    \ \ \ \ \ \ \ from this task to help out others.\\
    EE: Cool, that's good to know.  Have a great day!\\
    \midrule
    \textbf{3x3x5 ROLLOUTS (progression: 0.36; ER sentiment: 0.65; EE sentiment: 0.70)}\\
    \midrule
    ER: Well, it's not an international non-governmental organization, but they help children in war zones like Syria.  Children are our future and \\
    \ \ \ \ \ \ \ we need to help out where we can.\\
    EE: Ah, I see!  I will look into this.  Does it have any good ratings on Charity Navigator?\\
    ER: I have no idea, but I would be willing to look into them.  They're very trustworthy.\\
    EE: \textcolor{red}{They are very reputable. I will research them later and donate. Thanks!}\\
    ER: I'm glad to hear that.  This is something I would like to be passionate about.\\
    EE: Yes, I really love helping out!  I think we need to take care of our country's children.  And if you would like to know more about a charity \\
    \ \ \ \ \ \ \ you might like to research more:  URL\\
    ER: I would like to hear a little more about this charity from you.\\
    EE: Well, I will look that up.  This is one I really enjoy working on.  I've heard of some of the kids who are in need, but this is one I'm very \\
    \ \ \ \ \ \ \ passionate about.  Do you have any other charities you've worked on?\\
    ER: I'm going to look at some websites right now.  It's called Childrens International.  You can research them.  They are great.  They promote the \\
    \ \ \ \ \ \ \ rights of children in these situations, especially those that are separated from their parents or families.  Have you donated to them?\\
    EE: Not yet, but I think I will sometime in the future.\\
    \bottomrule
    \end{tabular}
\end{table*}

\begin{table*}[t]
    \vskip -4.10in
    \scriptsize
    \caption{Rollout Self-Play Example 3: \textbf{No RO} vs. \textbf{3x3x5} (dialogue \#98; generation seed: 589)}
    \label{response_generation_example_3}
    \begin{tabular}{l}
    \toprule
    \textbf{CONTEXT:} \\
    \midrule
    ER: Hi, do you usually donate to charities\\
    EE: I DO\\
    ER: great! You're such a nice person. What charities do you usually donate to?\\
    EE: To church, I return tithes and I also do the offerings\\
    ER: great! Well, have you ever heard about Save the Children charity?\\
    EE: no, can you tell me more about it?\\
    ER: it's an international non-governemntal organization that promotes children's rights, provides relief and helps support children in developing countries\\
    EE: nice! I have a child so i'm all about helping kids\\
    ER: would you donate to this charity? your donation will make such a huge impact for some children\\
    EE: I would like to know more before i donate. How do i know if it's a legit organization\\
    \midrule
    \textbf{NO ROLLOUTS (progression: -2.44; ER sentiment: 0.10; EE sentiment: 0.41)}\\
    \midrule
    ER: if you would like to donate, you can visit their site:  URL\\
    EE: i dont want to donate, if i did however, how much would i donate?\\
    ER: if you have to donate, you would have to submit a donation agreement, i dont think you can donate right now\\
    EE: ok, how do i submit it?\\
    ER: You can't donate until you check out their website.\\
    EE: ok thanks for telling me, i will check it out\\
    ER: It is a charity that focuses on children in the United States\\
    EE: ok thanks for the information, if i ever get to a child's situation, i will donate\\
    ER: You're very welcome\\
    EE: ok, have a nice day\\
    \midrule
    \textbf{3x3x5 ROLLOUTS (progression: 1.14; ER sentiment: 0.36; EE sentiment: 0.52)}\\
    \midrule
    ER: I will send you a copy of the actual donation and you can make a donation from there.\\
    EE: How do i donate?\\
    ER: It's in your task payment. You can do that right now.\\
    EE: How do i donate?\\
    ER: You don't need to donate at this point, but I'll give you a heads up on the website:  URL\\
    EE: Ok, i'll go check it out now. Thank you very much!\\
    ER: You're welcome, but I have to ask, Would you be willing to donate \$2, to the Save the Children charity?\\
    EE: \textcolor{red}{Ok. I will probably just donate \$2.}\\
    ER: That is great. Thank you so much for your donation!\\
    EE: You're very welcome. Thank you, have a nice one!\\
    \bottomrule
    \end{tabular}
\end{table*}

\end{document}